%% file: paper.tex
\documentclass{article}

\usepackage{microtype}
\usepackage{graphicx}
\usepackage{booktabs} % for professional tables

\usepackage{hyperref}

\usepackage[accepted]{icml2020}

\icmltitlerunning{On Value Functions and the Agent-Environment Boundary}

\usepackage{natbib}
\usepackage{subcaption}

\input{def}

\newcount\Comments  % 0 suppresses notes to selves in text
\definecolor{darkgreen}{rgb}{0,0.5,0}
\definecolor{darkred}{rgb}{0.7,0,0}
\definecolor{teal}{rgb}{0.3,0.8,0.8}
\newcommand{\kibitz}[2]{\ifnum\Comments=1\textcolor{#1}{#2}\fi}

\newcommand{\para}[1]{\vspace*{1pt} \noindent \textbf{#1}~}

\begin{document}
	
\twocolumn[
\icmltitle{On Value Functions and the Agent-Environment Boundary}

\begin{icmlauthorlist}
\icmlauthor{Nan Jiang}{uiuc}
\end{icmlauthorlist}
	
\icmlaffiliation{uiuc}{University of Illinois at Urbana-Champaign}

\icmlcorrespondingauthor{Nan Jiang}{nanjiang@illinois.edu}
	
\icmlkeywords{reinforcement learning}

\vskip 0.3in
]
	
\printAffiliationsAndNotice{}  % leave blank if no need to mention equal contribution

\begin{abstract}
When function approximation is deployed in reinforcement learning (RL), the same problem may be formulated in different ways, often by treating a pre-processing step as a part of the environment or as part of the agent. As a consequence, fundamental concepts in RL, such as (optimal) value functions, are not uniquely defined as they depend on where we draw this \emph{agent-environment boundary}. This causes further problems in theoretical analyses that provide optimality guarantees, as the same analysis may yield different bounds in equivalent formulations of the same problem. We address this issue via a simple and novel \emph{\bi} analysis of Fitted Q-Iteration, a representative RL algorithm, where the assumptions and the guarantees are invariant to the choice of boundary. 
We also discuss closely related issues on state resetting, deterministic vs stochastic systems, imitation learning, and the verifiability of theoretical assumptions from data. 
\end{abstract}

\section{Introduction} \label{sec:intro}
A large part of RL theory---including that on function approximation---% algorithms---including those designed to handle large state spaces with function approximation---
is built on 
%we characterize and understand algorithms via 
mathematical concepts established in the Markov Decision Process (MDP) literature \citep{puterman1994markov}, 
%Our theoretical understanding of reinforcement learning (RL) algorithms is deeply rooted in the mathematical concepts established in the Markov Decision Process (MDP) literature \cite{puterman1994markov}, 
such as the optimal state- and $Q$-value functions ($V^\star$ and $Q^\star$) and their policy-specific counterparts ($V^\pi$ and $Q^\pi$). These functions operate on the state (and action) of the MDP, and classical results tell us that they are always uniquely and well defined. % through the Bellman optimality equations and Bellman equations for policy evaluation.  

%It is common for a researcher to start the design of a  new algorithm by saying, ``let's use a function approximator (e.g., a neural network) to approximate $Q^\star$ of this environment.''

Are they really well defined?

\begin{figure*}[ht]
	\centering
	\includegraphics[width=\textwidth]{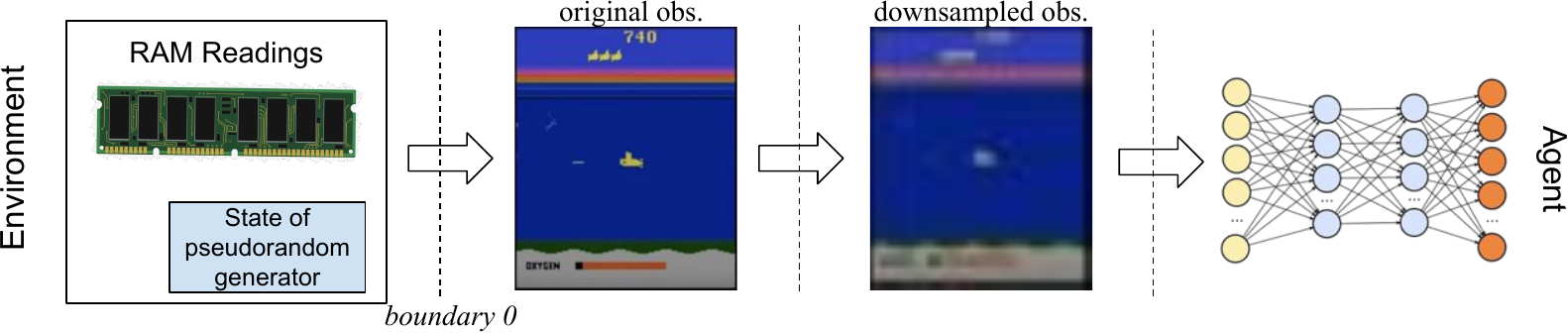}
	\caption{Illustration of the agent-environment boundaries. Strickly speaking, the environments defined by the intermediate boundaries are partially observable, but we can view them as MDPs over histories (of actions and observations defined by the boundary). Therefore, partial observability has little to do with our concerns, and we stick to MDP terminologies in the main text and do not invoke POMDP concepts for simplicity and clarity.  \label{fig:boundary}}
\end{figure*}

Consider the following scenario, depicted in Figure~\ref{fig:boundary}. In a standard ALE benchmark \citep{bellemare2013arcade},  raw-pixel screens are produced as states (or strictly speaking, observations\footnote{%RL environments are partially observable in general, and even if the original environment is Markov, partial observability can arise due to lossy processing of states.  
%In Atari, it is common to include past frames in the state representation to resolve partial observability. % which is omitted in Figure~\ref{fig:boundary}. 
In most of the paper we stick to MDP terminologies for simplicity, but the discussions and implications also apply to partially observable systems (which can always be viewed as MDPs over histories); see the caption of Figure~\ref{fig:boundary} for more details.}), 
and the agent feeds the state into a neural net to predict $Q^\star$. Since the original game screen has a high resolution, %resulting in high-dimensional states, 
it is common in practice to downsample the  screen as a pre-processing step \citep{mnih2015human}.

There are two equivalent views of this scenario: We can view the pre-processing step as part of the environment, or as part of the agent. 
%To be more formal, let $s$ denote the original game screen, and $\phi$ denote the pre-processing procedure that produces the low-dimensional screen $\phi(s)$. In the first view, the environment is an MDP with state $s$, denoted as $M$, and in the second view, the environment is an MDP with state $\phi(s)$, denoted as $M_{\phi}$. These two views only differ in the subjective choice of the \emph{agent-environment boundary} with no real consequences. However, if $\phi$ is a lossy process (in general it will be), the $Q^\star$ of the two MDPs can be different.\footnote{The two functions operate on different state spaces, but can be compared naturally after lifting \cite{}, in the sense that $Q^\star_{M}(\phi(\cdot), \cdot) \ne Q^\star_{M_\phi}(\cdot, \cdot)$, where the subscript denotes the MDP w.r.t.~the $Q^\star$ is defined.} 
Depending on where we draw this \emph{agent-environment boundary}, %---which is subjective and has no real consequences---
$Q^\star$ will be different in general. %\footnote{For concreteness, we provide a minimal example of \bdc in Appendix~\ref{app:concrete}.} 
It should also be obvious that there may exist many choices of the boundary (e.g., ``\emph{boundary 0}'' in Figure~\ref{fig:boundary}), some of which we may be even not aware of. When we design an algorithm to learn $Q^\star$, %(that is, the function approximator \emph{realizes} $Q^\star$), 
which $Q^\star$ are we talking about? 

The good news is that many existing algorithms are \emph{\bxi}, that is, once the function approximation scheme is \textbf{fixed}, the boundary is only a subjective choice and does not affect the behavior of the algorithm. The bad news is that many existing analyses\footnote{There are different kinds of theoretical analyses in RL (e.g., convergence analysis). In this paper we focus on analyses that provide near-optimality guarantees.} are \emph{\bxd}, as they make assumptions that may either hold or fail in the same problem depending on the choice of the boundary: for example, in the analyses of approximate value iteration algorithms, it is common to assume that $Q^\star$ can be represented by the function approximator (``\emph{realizability}''), and that the function space is closed under Bellman update \citep[low ``\emph{inherent Bellman error}'', ][]{szepesvari2005finite, antos2008learning}, both of which are \bd assumptions. Such a gap between the mathematical theory and the reality also leads to further consequences, such as the theoretical assumptions being fundamentally unverifiable from naturally generated data. 

In this paper we systematically study the \bdc of RL theory. We ground our discussions in a simple and novel \emph{\bi} analysis of Fitted Q-Iteration~\cite{ernst2005tree}, %a representative batch-RL algorithm, 
in which the correctness of the assumptions and the guarantees do not change with the subjective choice of the boundary (Sections~\ref{sec:cb2} and \ref{sec:fqi}). Within this analysis, we give up on the classical notions of value functions or even the (state-wise) Bellman equation, and replace them with weaker conditions that are \bxi. % and that naturally come with improved verifiability. 
We also discuss closely related issues regarding state resetting, deterministic versus stochastic systems, imitation learning, and the verifiability of theoretical assumptions from data (Section~\ref{sec:conclusion}). The implications drawn from our results suggest novel ways of thinking about states, value functions, and optimality in RL.

\section{Preliminaries} \label{sec:prelim}
\para{Markov Decision Processes}~
An infinite-horizon discounted MDP $M$ is specified by $(\Scal, \Acal, P, R, \gamma, d_0)$, where $\Scal$ is the finite state space,\footnote{For the ease of exposition we assume finite $\Scal$, but its cardinality can be arbitrarily large.} $\Acal$ is the finite action space, $P: \Scal\times\Acal\to\Delta(\Scal)$ is the transition function,\footnote{$\Delta(\cdot)$ is the probability simplex.} $R: \Scal\times\Acal\to\Delta([0, \Rmax])$ is the reward function, $\gamma \in [0, 1)$ is the discount factor, and $d_0 \in \Delta(\Scal)$ is the initial state distribution. %, which is common in e.g.., episodic tasks \citep{williams1992simple, kakade2002approximately, sutton2018reinforcement}. 

A (stationary and deterministic) policy $\pi: \Scal \to \Acal$ specifies a decision-making strategy, and induces a distribution over random trajectories: $s_1 \sim d_0$, $a_1 \sim \pi$, $r_1 \sim R(s_1, a_1)$, $s_2 \sim P(s_1, a_1)$, $a_2 \sim \pi$, \ldots, where $a_t \sim \pi$ is short for $a_t = \pi(s_t)$. In later analyses, we will also consider stochastic policies $\pi: \Scal\to\Delta(\Acal)$ and non-stationary policies formed by concatenating a sequence of stationary ones.

The performance of a policy $\pi$ is measured by its expected discounted return (or value):\footnote{It is important that the performance of a policy is measured under the initial state distribution. See Appendix~\ref{app:dbrm} for further discussions.}
$$\textstyle
v^\pi:= \EE[\sum_{t=1}^\infty \gamma^{t-1} r_t \cond s_1 \sim d_0, \, a_{1:\infty} \sim \pi].
$$
The value of a policy lies in the range of $[0, \Vmax]$ with $\Vmax = \Rmax/(1-\gamma)$. 
It will be useful to define the $Q$-value function of $\pi$:  
%\begin{align} \label{eq:Qpi}
$\textstyle
Q^\pi(s,a) := \EE[\sum_{t=1}^\infty \gamma^{t-1} r_t \cond s_1 =s, a_1 = a, \, a_{2:\infty} \sim \pi],
$ %\end{align}
and $d_t^\pi$, the distribution over state-action pairs induced at time step $t$: 
%\begin{align}\label{eq:sadist}
$d_t^\pi(s,a) := \Pr[s_t = s, a_t = a \cond s_1 \sim d_0, \, a_{1:\infty} \sim \pi].$ 
%\end{align}
Note that $d_1^\pi = d_0 \times \pi$, which means $s\sim d_0, a \sim \pi$.

The goal of the agent is to find a policy $\pi$ that maximizes $v^\pi$. In the infinite-horizon discounted setting, there always exists an optimal policy $\pi^\star$ that maximizes the expected discounted return for all states simultaneously (and hence also for $d_0$). Let $Q^\star$ be a shorthand for $Q^{\pi^\star}$. It is known that $\pi^\star$ is the greedy policy w.r.t.~$Q^\star$: For any $Q$-function $f$, let $\pi_f$ denote its greedy policy $(s \mapsto \argmax_{a\in\Acal}f(s,a))$, and we have $\pi^\star = \pi_{Q^\star}$. Furthermore, $Q^\star$ satisfies the Bellman equation: $Q^\star = \Tcal Q^\star$, where $\Tcal: \RR^{\Scal\times \Acal} \to \RR^{\Scal\times \Acal}$ is the Bellman optimality operator:
\begin{align*}
(\Tcal f)(s,a) = \EE_{r\sim R(s,a), s'\sim P(s,a)}[r + \gamma \max_{a'\in\Acal} f(s',a')].
\end{align*}

\para{Value-function Approximation} 
In complex problems with high-dimensional observations, function approximation is often deployed to generalize over the large state space. In this paper we take a learning-theoretic view of value-function approximation: We are given a function space $\Fcal\subset(\Scal\times\Acal \to [0, \Vmax])$, and for simplicity we assume $\Fcal$ is finite.\footnote{The only reason that we assume finite $\Fcal$ is for mathematical convenience in Theorem~\ref{thm:fstar}, and removing this assumption only has minor impact on our results. See further comments after Theorem~\ref{thm:fstar}'s proof in Appendix~\ref{app:fqi}.} The goal---stated in the classical, \bd fashion---is to identify a function $f\in\Fcal$ such that $f\approx Q^\star$, so that $\pi_f$ is a near-optimal policy. This naturally motivates a common assumption, known as \emph{realizability}, that %A typical assumption commonly found in literature is \emph{realizability}, that is, 
$Q^\star \in \Fcal$, which will be useful for later discussions. %In fact, most batch value-based RL algorithms require stronger assumptions on $\Fcal$ to succeed \cite{szepesvari2005finite, antos2008learning}, which we will introduce in Section~\ref{sec:fqi}.

For most of the paper we will be concerned with batch-mode value-function approximation, that is, the learner is passively given a dataset consisting of tuples $(s,a,r,s')$ and cannot directly interact with the environment. The implications in the exploration setting will be briefly discussed at the end of the paper. 

\para{Fitted Q-Iteration (FQI)}
FQI \citep{ernst2005tree, szepesvari2010algorithms} is a batch RL algorithm that solves a series of least-squared regression problems with $\Fcal$ to approximate each step of value iteration. It is also considered as the prototype for the popular DQN \citep{mnih2015human}, and often used as a representative of off-policy value-based RL algorithms in empirical studies \citep{fu2019diagnosing}. We defer a detailed description of the algorithm to Section~\ref{sec:fqi_intro}.

\section{Case Study: \BDC in Batch Contextual Bandit (CB) with Predictable Rewards} \label{sec:cb}

To give a concrete instance of how standard theory is \bxd, we consider the simplified setting of contextual bandits, which may be viewed as episodic MDPs with $\gamma=0$; here each trajectory only lasts for 1 step and the initial state distribution $d_0$ corresponds to the context distribution of the CB \cite{langford2008epoch}. In CB, FQI degrades to fitting the reward function via squared loss regression. In the rest of this section we will introduce the setting and the algorithm formally, and provide a minimal example to illustrate how difference choices of the boundaries can result in different bounds in the same problem.

%We warm-start with the simple problem of fitting a reward function from batch data in contextual bandits.\footnote{%We keep the notations consistent so that 
%In  Sections~\ref{sec:cb} and \ref{sec:fqi}, the same symbol carries the same (or similar) meaning. However, there are some inevitable differences and the reader should not confuse the two settings in general.} The analysis also applies straightforwardly to learning a policy-specific value-function $Q^\pi$ from Monte-Carlo rollouts and performing one-step policy improvement \citep{sutton2018reinforcement}. This section also provides important building blocks for Section~\ref{sec:fqi}, and the simplicity of the analysis allows us to thoroughly discuss the intuitions and the conceptual issues, leaving Section~\ref{sec:fqi} focused on the technical aspects.

\subsection{Setting} \label{sec:cb_setting}
Let $D=\{(s_i, a_i, r_i)\}$ be a dataset, where $s_i \sim d_0$, $a_i\sim \pi_b$ and $r_i \sim R(s_i,a_i)$. Here $\pi_b$ is a behavior policy with which we collect the data. Let $\mu$ denote the joint distribution over $(s,a)$, or $\mu := d_0 \times \pi_b$. %We are equipped with a function class $\Fcal \subset (\Scal\times\Acal\to [0, \Rmax])$, and 
For any $f\in\Fcal$, define the empirical squared loss
\begin{align} \label{eq:cb_loss} \textstyle
\Lcal_{D}(f) := \frac{1}{|D|}\sum_{i} (f(s_i,a_i) - r_i)^2,
\end{align}
and the population version $
\Lcal_{\mu}(f) := \EE_{D}[\Lcal_D(f)].
$ 
The algorithm fits a reward function by minimizing $\Lcal_D(\cdot)$, that is,
$
\hf := \argmin_{f\in\Fcal} \Lcal_D(f),
$ 
and outputs $\pi_\hf$. We are interested in providing a guarantee to the performance of this policy, that is, the expected reward obtained by executing $\pi_\hf$, 
$
v^{\pi_\hf} = \EE[r \cond \pi_\hf].
$
We will base our analyses on the following inequality:
\begin{align} \label{eq:eps}
\Lcal_{\mu}(\hf) -  \min_{f\in\Fcal} \Lcal_{\mu}(f) \le \epsilon.
\end{align}
In words, we assume that $\hf$ approximately minimizes the population loss. Such a bound can be obtained via a uniform convergence argument, where $\epsilon$ will depend on the sample size $|D|$ and the statistical complexity of $\Fcal$ (e.g., pseudo-dimension \citep{haussler1992decision}). We do not include this part as it is standard and orthogonal to the discussions in this paper, and rather focus on how to provide a guarantee on $v^{\pi_\hf}$ as a function $\epsilon$.

\subsection{Classical Assumptions and Guarantees}
We now review the classical assumptions in this problem for later references and comparisons.  The first assumption is that data is exploratory, often guaranteed by taking randomized actions in the data collection policy (or behavior policy $\pi_b$) and not starving any of the actions:

\begin{assumption}[$\pi_b$ is exploratory] \label{asm:cb_explore}
There exists a universal constant $C < +\infty$ s.t.,  $\forall s\in\Scal, a\in\Acal$, ~~
$
\pi_b(a | s) \ge 1/C.
$
\end{assumption}

Next is realizability as already discussed in Section~\ref{sec:prelim}.

\begin{assumption}[Realizability]  \label{asm:cb_realizable}
$Q^\star \in \Fcal$. In contextual bandits, $Q^\star(s,a) = \EE[r \,|\, s,a]$, $\forall (s,a)$, which is the reward function. % \in \Scal\times\Acal$.
\end{assumption}

With these two assumptions, the standard guarantee is the following:
\begin{theorem} \label{thm:cb_lt}
Under Assumptions~\ref{asm:cb_explore} and \ref{asm:cb_realizable},  \\
$
v^{\pi_\hf} \ge v^\star - 2\sqrt{C\epsilon}.
$
\end{theorem}
The proof is omitted as the theorem will be  subsumed by more general results. (It is a corollary of either Theorem~\ref{thm:cb_robust} or \ref{thm:cb}). Due to the straightforwardness of the analyses and that most CB literature focus on the online setting, we are unable to find an appropriate citation, but we believe readers familiar with the literature will agree that this is a natural setup and analysis for the batch setting. 

%Two comments before we move on: 
%\begin{compactenum}[\textbullet]\setlength{\itemsep}{5pt}
%\item While we consider exact realizability for simplicity, it is possible to allow an approximation error and state a guarantee that degrades gracefully with the violation of the assumption. Such an extension is routine and we omit it for readability. 
%\item We do not provide the classical analysis here, and simply note that the guarantee is the same as Theorem~\ref{thm:cb} when Assumptions~\ref{asm:cb_explore} and \ref{asm:cb_realizable} hold.
%\end{compactenum}

\subsection{Illustration of \BDC}\label{sec:example}
\begin{figure}[t]
	\begin{subfigure}{0.26\textwidth}
		\centering
		$Q(s_A) = Q(s_B) = 0.5$
		\includegraphics[scale=1]{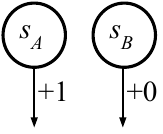}
		\caption{\label{fig:cb1}}
	\end{subfigure}
	\begin{subfigure}{0.2\textwidth}
		\centering
		$Q'(s) = 0.5$
		\includegraphics[scale=1]{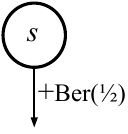}
		\caption{\label{fig:cb2}}
	\end{subfigure}
	\caption{\label{fig:cb} Illustration of two different formulations of the same problem when function approximation is deployed. \textbf{(a)} A contextual bandit with two contexts, $s_A$ and $s_B$, which appear with equal probabilities, i.e., $d_0(s_A) = d_0(s_B)=0.5$. The only action yields $+1$ reward and $+0$ reward in $s_A$ and $s_B$, respectively. The function approximator contains only 1 function $Q(s_A) = Q(s_B) = 0.5$ (action omitted since there is only 1 action). \textbf{(b)} A contextual bandit with one context $s$. The only action available yields a Bernoulli distributed stochastic reward. The function approximator contains only 1 function $Q'(s)=0.5$.}
\end{figure}
We are now ready to illustrate the \bdc of Theorem~\ref{thm:cb_lt} using a minimal example. 

\para{Weaker Sense} We start with a weaker sense of \bdc, that the assumptions which Theorem~\ref{thm:cb_lt} rely on may hold under one boundary and fail under another. Consider the two contextual bandit problems in Figure~\ref{fig:cb}, each equipped with a function class for modeling the reward function. It should be obvious that, from the viewpoint of the learning algorithm described in Section~\ref{sec:cb_setting}, the two problems are fundamentally indistinguishable, as \figref{cb1} becomes \figref{cb2} when the learner fails to distinguish between $s_A$ and $s_B$. % as long as every function in the candidate class assigns equal values to $s_A$ and $s_B$ in \figref{cb1}. %, it is as if there were a pre-processing step that aggregates $s_A$ and $s_B$ together, which is equivalent to \figref{cb2}. 
However, Assumption~\ref{asm:cb_realizable} holds in \figref{cb2} but fail in \figref{cb1}, as the reward function for \figref{cb1} is $Q^\star(s_A) = 1, Q^\star(s_B) = 0$, which is not in the class. 

\para{Stronger Sense} Familiar readers may argue that the weaker sense of \bdc is not necessarily a problem: although Theorem~\ref{thm:cb_lt} becomes vacuous when we change the boundary, we can still provide meaningful guarantees by considering the approximation errors in the violation of Assumption~\ref{asm:cb_realizable}, like in the following result. 
\begin{theorem}[Robust version of Theorem~\ref{thm:cb_lt}]\label{thm:cb_robust}
Define
$
\eapx:= \inf_{f\in\Fcal} \|f - Q^\star\|_{\mu}^2.
$ 
Under Assumption~\ref{asm:cb_explore} alone, we have~~
$
v^{\pi_\hf} \ge v^\star - 2\sqrt{C(\epsilon + \eapx)}.
$
\end{theorem}
When $\eapx=0$ we recover Theorem~\ref{thm:cb_lt}. As $\eapx$ increases, the guarantee degrades gradually. % as Assumption~\ref{asm:cb_realizable} is violated more severely. 
Unfortunately, such a robustness result does not save the classical analysis from the \bdc issue. In fact, the bounds are simply different when we apply Theorem~\ref{thm:cb_robust} in \figref{cb1} and \figref{cb2}: %, and ideally we would like to see the same guarantees. However, 
when $\epsilon=0$, the value guaranteed for \figref{cb2} is $0.5$, while that for \figref{cb1} is $0.5 - 2 \sqrt{0.25} = -0.5$  (here $C=1$), which is much looser!\footnote{In fact, it is not hard to see that the bound will be lower than $0.5$ if any approximation error is measured. That is, even if a different analysis is considered (we believe our bound is tight as a general analysis), as long as the bound penalizes the violation of Assumption~\ref{asm:cb_explore}, it will provide worse guarantee in \figref{cb2} than in \figref{cb1}.} 

Of course, no loss should be incurred anyway as there is only 1 action in Figure~\ref{fig:cb}, and this example is only intended to illustrate the inconsistency of the classical analyses when they are applied to different formulations of the same problem. While problem-specific and ad hoc fixes may be plausible, %we are not aware of any approach that completely fixes this issue. W
what we present in this paper is a general strategy that provides guarantees competitive to the classical analyses under \emph{any} boundary, together with interesting implications that require us to rethink the meaning of states, value functions, and optimality in RL. 

\section{\BI Analysis for Batch CB} \label{sec:cb2}

\subsection{A Sufficient Condition for \BIC} 
Before we start the analysis, we first show that the algorithm itself is \bxi, leading to a sufficient condition for judging the \bic of analyses. The concept central to the algorithm is the squared loss $\Lcal_D(f)$. Although $\Lcal_D(f)$ is defined using $(s,a)$, the definition refers to $(s,a)$ exclusively through the evaluation of $f\in\Fcal$ on $(s,a)$, taking expectation over a naturally generated dataset $D$. %\footnote{Here we use ``naturally generated'' to contrast state resetting operations discussed in Section~\ref{sec:dilemma}.} 
The data points are generated by an objective procedure (collecting data with policy $\pi_b$), and on every data point $(s,a,r)$, $f(s,a)$ is the same scalar regardless of the boundary, hence the algorithmic procedure is \bi. 

Inspired by this, we provide the following sufficient condition for \bi analyses:

\begin{claim} \label{clm:bi}
An analysis is \bxi if the assumptions and the optimal value are defined in a way that accesses states and actions exclusively through evaluations of functions in $\Fcal$, with plain expectations (either empirical or population) over natural data distributions.
\end{claim}

A number of pitfalls need to be avoided in the specification of such a condition: 
\begin{compactenum}[\textbullet]
\setlength{\itemsep}{5pt}
\item Restricting the functions to $\Fcal$ is important, as one can define conditional expectations (on a single state) through plain expectations via the use of state indicator functions (or the dirac delta functions for continuous state spaces). 
\item Similarly, we need to restrict the set of distributions to those natural ones (formalized later in Definition~\ref{def:admissible}) to prevent expectations on point masses. (See Appendix~\ref{app:dbrm} for further discussions.)
\item Besides the assumptions, the very notion of optimality also needs to be taken care of, as $v^\star := \EE_{s \sim d_0}[V^\star(s)]$ (the usual notion of optimal value) is also a \bd quantity. (See Section~\ref{sec:bd_list} for further discussions.)
%\item %which are closely tied to the initial state distribution $d_0$. For readers who wonder why this has to be the case, see further discussions in Appendix~\ref{app:dbrm}.
\end{compactenum}

That said, this condition is not perfectly rigorous, as we find it difficult to make it mathematically strict without being verbose and/or restrictive. Regardless, we believe it conveys the right intuitions and can serve as a useful guideline for judging the \bic of a theory. 
Furthermore, the condition provides us with significant mathematical convenience: as long as the condition is satisfied, we can analyze an algorithm under \emph{any} boundary, allowing us to use the standard MDP formulations and all the objects defined therein (states, actions, their distributions, etc.). % despite that they may be \bd.

\subsection{Assumptions and Analysis} \label{sec:cb_bi}
%We present the \bi counterparts of Assumptions~\ref{asm:cb_explore} and \ref{asm:cb_realizable}, and analyze the algorith. 
\para{Additional Notations} For any $f: \RR^{\Scal\times\Acal}$ and any distribution $\nu\in\Delta(\Scal\times\Acal)$, define $\|f\|_{2, \nu}^2 := \EE_{(s,a)\sim \nu}[f(s,a)^2]$. Since we will only be concerned with weighted $\ell_2$ norm, we will abbreviate $\|f\|_{2, \nu}^2$ as $\|f\|_{ \nu}^2$ throughout the paper. To improve readability we will often omit ``$r \sim R(s,a)$'' when an expectation involves $r$ (and ``$s' \sim P(s,a)$'' for Section~\ref{sec:fqi}), and use $\EE_{\nu}[f]$ as a shorthand for $\EE_{(s,a)\sim \nu}[f(s,a)]$.

\begin{definition}[Admissible distributions (bandit)] \label{def:cb_admissible} 
Given a contextual bandit problem and a space of candidate reward functions $\Fcal$, we call $\{d_0 \times \pi_f : f\in\Fcal\}$ the space of \emph{admissible distributions}.
\end{definition}

\begin{assumption} \label{asm:bi_cb_explore}
There exists a universal constant $C < +\infty$ such that, for any $f, f' \in \Fcal$ and any admissible $\nu$, ~
$
\|f- f'\|_{\mu}^2 \le C \cdot\|f - f'\|_{\nu}^2.
$
\end{assumption}
Assumption~\ref{asm:bi_cb_explore} is a direct consequence of Assumption~\ref{asm:cb_explore}, and any $C$ that satisfies the latter also works for the former. %as $C$ is an upper bound on the $\ell_\infty$ norm of the importance ratio between $\nu$ and $\mu$.
The proof is elementary and omitted. 

\begin{assumption}\label{asm:bi_cb_realizable}
There exists $f^\star\in\Fcal$, such that %$f^\star$ is a valid reward function w.r.t.~all admissible distributions, 
for all admissible $\nu$, 
\begin{align} \label{eq:reward}
\EE_{\nu}[f^\star] = \EE_{(s,a)\sim \nu}[r],
\end{align}
and for any $f'\in\Fcal$,
\begin{align} \label{eq:sqloss}
\Lcal_{\mu}(f') - \Lcal_{\mu}(f^\star) = \|f' - f^\star\|_\mu^2.
\end{align}
We say that such an $f^\star$ is a valid reward function of the CB w.r.t.~$\Fcal$.
\end{assumption}
%This assumption is of high importance to this section. 
Assumption~\ref{asm:bi_cb_realizable} is implied by Assumption~\ref{asm:cb_realizable}, as $f^\star = Q^\star$ satisfies both Eq.\eqref{eq:reward} and Eq.\eqref{eq:sqloss}: Eq.\eqref{eq:reward} can be obtained from $Q^\star(s,a) = \EE[r| s,a]$ by taking the expectation of both sides w.r.t.~$\nu$. 
Eq.\eqref{eq:sqloss} is the standard bias-variance decomposition for squared loss regression when $f^\star$ is the Bayes-optimal regressor.\footnote{It is easy to allow an approximation error in Eq.\eqref{eq:reward} and/or \eqref{eq:sqloss}. For example, one can measure the violation of Eq.\eqref{eq:reward} by $\inf_{f\in\Fcal} \sup_{\nu} |\EE_{(s,a)\sim \nu}[f - r]|$, and such errors can be easily incorporated in our later analysis.} 
Eq.\eqref{eq:reward} guarantees that $f^\star$ still bears the semantics of reward, although no longer in a point-wise manner. Eq.\eqref{eq:sqloss} guarantees that $f^\star$ can be reliably identified through squared loss minimization, which is specialized to and required by the squared loss minimization approaches. In fact, we provide a counter-example in Appendix~\ref{app:sqloss} showing that dropping Eq.\eqref{eq:sqloss} can result in the failure of the algorithm. %, and also discuss other learning settings where this assumption is not needed. 

Now we are ready to state the main theorem of this section, whose proof %. All proofs for  Sections~\ref{sec:cb} and \ref{sec:fqi} 
can be found in Appendix~\ref{app:fqi}. (Also note that Theorem~\ref{thm:cb_lt} is a direct corollary of Theorem~\ref{thm:cb}.)

\begin{comment}
\begin{fact}
Assumptions~\ref{asm:bi_cb_explore} and  
\ref{asm:bi_cb_realizable} are direct consequences of  Assumption~\ref{asm:cb_explore} and \ref{asm:cb_realizable}, respectively.
\end{fact}

\begin{fact}
Assumptions~\ref{asm:bi_cb_explore} and  
\ref{asm:bi_cb_realizable} are \bi.
\end{fact}
\end{comment}

\begin{theorem} \label{thm:cb}
Under Assumptions~\ref{asm:bi_cb_explore} and \ref{asm:bi_cb_realizable}, for any valid $f^\star$, we have
$
v^{\pi_\hf} \ge v^{\pi_{f^\star}} - 2\sqrt{C\epsilon}.
$
\end{theorem}

\para{Comparison to Theorem~\ref{thm:cb_robust}} We  compare our \bi analyses with standard \bd analyses both in the example of Figure~\ref{fig:cb} and more generally. Observe that in \figref{cb1}, $Q(s_A) = Q(s_B) = 0.5$ is considered valid by Assumption~\ref{asm:bi_cb_realizable}, so we can directly compete with $v^{\pi_Q} = 0.5$. The bound is tight when $\epsilon=0$, which is in sharp contrast to the loose bound given by Theorem~\ref{thm:cb_robust} (see Section~\ref{sec:example}). 

More generally, if Assumption~\ref{asm:cb_realizable} holds under any boundary, Assumption~\ref{asm:bi_cb_realizable} holds regardless of how the problem is formulated. Consequently,  when $\epsilon=0$ the \bi bound guarantees an optimal value which cannot be exceeded by any policy in $\{\pi_f: f\in\Fcal\}$, while the \bd bounds are generally loose. It is actually possible to push this further and prove the superiority of \bi bounds in more general conditions:
\begin{proposition} \label{prop:bi_better}
When Assumptions~\ref{asm:bi_cb_explore} and \ref{asm:bi_cb_realizable} hold, Theorem~\ref{thm:cb} is always as tight as Theorem~\ref{thm:cb_robust} as long as $\epsilon \le \eapx/2$. 
\end{proposition} 
The proof is deferred to Appendix~\ref{app:fqi}. The proposition states that the \bi guarantee in Theorem~\ref{thm:cb} is always as tight as the robust result given in Theorem~\ref{thm:cb_robust}, as long as the best function in class measures a \bd approximation error $\eapx$ that is greater than $\epsilon$, the error due to finite sample effect and/or inexact optimization (Eq.\eqref{eq:eps}). %In the regime of $\epsilon > \eapx/2$, even though the \bi guarantee might not be exactly tighter than the \bd guarantee, the suboptimality gaps will be similar as both of them will be dominated by $\epsilon$. 

Note that the above result only applies when Assumption~\ref{asm:bi_cb_realizable} holds. We do expect that the superiority of \bi analyses should hold in more general situations---at least in some approximate sense---even when Assumption~\ref{asm:bi_cb_realizable} fails (in which case we will need a robust version of Theorem~\ref{thm:cb}), but the analyses become much more involved and it is unclear if a clean analysis and apple-to-apple comparison can be done, so we leave a deeper investigation to future work. 

%let $(M, \Fcal)$ be a pair of MDP environment and function approximator such that $Q_M^\star \in \Fcal$ (e.g., Figure~\ref{fig:cb2}). (Here we use subscript to clarify the MDP w.r.t.~$Q^\star$ is computed.) Now $(M, \Fcal)$ may be equivalent to some other pair $(M', \Fcal')$ (e.g., \figref{cb1}). Suppose Assumption~\ref{asm:cb_realizable} is satisfied under \emph{some} formulation, say, $(M, \Fcal)$, Theorem~\ref{thm:cb} will be tight for $\epsilon=0$, as $Q^\star \in \Fcal$ is valid and we can achieve the optimal value $v^\star$ of this problem. 

\section{Case Study: Fitted Q-Iteration}
\label{sec:fqi}
With the preparation in Section~\ref{sec:cb2}, we now analyze the FQI algorithm, and prove the counterparts of Theorems~\ref{thm:cb_lt} and \ref{thm:cb} for FQI. 

The major technical difficulty in this section is that the \bi counterpart of $Q^\star$ is not automatically defined in the \bi analysis, and we have to establish it from the assumptions without using familiar concepts in MDP literature (e.g., we cannot use the fact that $\Tcal$ is a $\gamma$-contraction under $\ell_\infty$ in the \bi analysis). 

\subsection{Setting and Algorithm} \label{sec:fqi_intro}
To highlight the differences between \bd and \bi analyses, we adopt a simplified setting assuming i.i.d.~data. Interested readers can consult prior works for more general analyses on $\beta$-mixing data \citep[e.g.,][]{antos2008learning}. 

Let $D=\{(s_i, a_i, r_i, s_i')\}$ be a dataset, where $(s_i,a_i) \sim \mu$, $r_i \sim R(s_i,a_i)$, and $s_i'\sim P(s_i,a_i)$. %The algorithm is given a function class $\Fcal \subset (\Scal\times\Acal\to [0, \Vmax])$, and
For any $f, f'\in\Fcal$, define the empirical squared loss
$$ \textstyle
\Lcal_{D}(f; f') := \frac{1}{|D|}\sum_{i} (f(s_i,a_i) - r_i- \gamma \max_{a'\in\Acal} f'(s_i',a'))^2,
$$
and the population version $\Lcal_{\mu}(f;f') := \EE_{D}[\Lcal_D(f;f')]$. 
The algorithm initializes $f_1$ arbitrarily, and 
$$ \textstyle
f_i := \argmin_{f\in\Fcal} \Lcal_D(f; f_{i-1}), ~~~ \text{for~ $i\ge 2$.}
$$
The algorithm repeats this for some $k$ iterations and outputs $\pi_{\fh}$. We are interested in providing a guarantee to the performance of this policy. %, that is, the expected discounted return obtained by executing $\pi_\hf$, 
%~$
%v^{\pi_\hf} := \EE[\sum_{t=1}^\infty \gamma^{t-1} r_t \cond \pi_\hf].
%$

\subsection{Classical Assumptions}

Similar to the CB case, there will be two assumptions, one that requires the data to be exploratory, and one that requires $\Fcal$ to satisify certain representation conditions.

\begin{definition}[Admissible distributions (MDP)] \label{def:admissible} 
A state-action distribution is admissible if it takes the form of $d_t^\pi$ for any $t$, and any non-stationary policy $\pi$ formed by choosing a policy for each time step from $\{\pi_f: f\in\Fcal\} \bigcup \{\pi_{f,f'}: f, f'\in\Fcal\}$, where $\pi_{f,f'}(s) := \argmax_{a\in\Acal} \max(f(s,a), f'(s,a))$. %The probability mixture of such distributions is also admissible.
\end{definition}

\begin{assumption}[$\mu$ is exploratory] \label{asm:explore}
There exists a universal constant $C < +\infty$ such that for any admissible $\nu$,
$\max_{s,a}\frac{\nu(s,a)}{\mu(s,a)} \le C$. 
\end{assumption}
This guarantees that $\mu$ well covers all admissible distributions. The upper bound $C$ is known as the \emph{concentrability coefficient} \citep{munos2003error}, and here we use the simplified version from a recent analysis by \citet{chen2019information}. See \citet{farahmand2010error} for a more fine-grained characterization of this quantity. 

\begin{assumption}[No inherent Bellman error]  \label{asm:complete}
$\forall f \in \Fcal$, $\Tcal f\in\Fcal$.
\end{assumption}
This assumption states that $\Fcal$ is closed under the Bellman update operator $\Tcal$. It automatically implies $Q^\star \in \Fcal$ (for finite $\Fcal$) hence is stronger than realizability, but replacing this assumption with realizability can cause FQI to diverge \citep{van1994feature, gordon1995stable, tsitsiklis1997analysis} or suffer from exponential sample complexity \citep{dann2018oracle}. We refer the readers to \citet{chen2019information} for further discussions on the necessity of this assumption.

\para{\Bd Guarantee} Just as Theorem~\ref{thm:cb_lt} is a direct corollary of and immediately subsumed by Theorem~\ref{thm:cb}, the \bd guarantee for FQI under Assumptions~\ref{asm:explore} and \ref{asm:complete} will be subsumed later by the \bi guarantee in Theorem~\ref{thm:fqi} (which the readers can verify by referring to Theorem 2 of \citet{chen2019information} and its proof), so we do not present the result separately here. 

%It is also possible to relax the assumption and allow an approximation error in the form of $\sup_{f\in\Fcal} \inf_{f' \in \Fcal} \|f' - \Tcal f\|$, known as the \emph{inherent Bellman error} \citep{munos2008finite}. Again we do not consider this extension, and incorporating it in our analysis is straightforward.

\subsection{\BI Assumptions} \label{sec:bi_fqi_asm}
We give the \bi counterparts of Assumptions~\ref{asm:explore} and \ref{asm:complete}.

\begin{assumption} \label{asm:bi_explore}
There exists a universal constant $C < +\infty$ such that, for any $f, f' \in \Fcal$ and any admissible state-action distribution $\nu$, ~~~
$
\|f - f'\|_{\nu}^2 \le C \|f - f'\|_{\mu}^2.
$
\end{assumption}

\begin{assumption}\label{asm:bi_close}
$\forall f\in\Fcal$, there exists $g \in \Fcal$ 
such that for all admissible $\nu$, 
\begin{align} \label{eq:fqi_reward}
\EE_{\nu}[g] = \EE_{(s,a)\sim \nu}[r + \gamma \max_{a'\in \Acal}f(s',a')],
\end{align}
and for any $f'\in\Fcal$,
\begin{align} \label{eq:fqi_sqloss}
\Lcal_{\mu}(f'; f) - \Lcal_{\mu}(g; f) = \|f' - g\|_\mu^2.
\end{align}
Define $\Bcal$ as the operator that maps $f$ to an arbitrary (but systematically chosen) $g$ that satisfies the above conditions.
\end{assumption}
Assumption~\ref{asm:bi_close} states that for every $f\in\Fcal$, we can define a contextual bandit problem with random reward $r + \gamma \max_{a'\in\Acal} f(s', a')$, and there exists $g\in\Fcal$ that is a valid reward function for this problem (Assumption~\ref{asm:bi_cb_realizable}). In the classical definitions, the true reward function for this problem is $\Tcal f$, so our $\Bcal$ operator can be viewed as the \bi version of $\Tcal$.

\subsection{\BI Analysis}

In Section~\ref{sec:cb} for contextual bandits,  $f^\star$ is defined directly in the assumptions, and we use it to define the optimal value in Theorem~\ref{thm:cb}. In Assumptions~\ref{asm:bi_explore} and \ref{asm:bi_close}, however, no counterpart of $Q^\star$ is defined. How do we even express the optimal value that we compete with?

We resolve this difficulty by relying on the $\Bcal$ operator defined in Assumption~\ref{asm:bi_close}. Recall that in classical analyses, $Q^\star$ can be defined as the fixed point of $\Tcal$, so we define $f^\star$ similarly through $\Bcal$. 

\begin{theorem} \label{thm:fstar}
Under Assumption \ref{asm:bi_close}, 
there exists $f^\star \in \Fcal$ s.t.~$\|\Bcal f^\star - f^\star\|_{\nu} = 0$ for any admissible $\nu$.
\end{theorem}

The key to proving Theorem~\ref{thm:fstar} is to show a $\gamma$-contraction-like property of $\Bcal$, formalized in Lemma~\ref{lem:contraction}. %Note that this is different from the classical $\gamma$-contraction of $\Tcal$, which holds for $\ell_\infty$ norm, i.e.,  
\begin{lemma}[\Bi version of $\gamma$-contraction] 
\label{lem:contraction}
Under Assumption~\ref{asm:bi_close}, for any admissible $\nu$, 
$\forall f, f' \in \Fcal$, let $\pi_{f,f'}(s) := \argmax_{a\in\Acal} \max(f(s,a), f'(s,a))$, and $P(\nu)$ denote the distribution of $s'$ generated as $(s,a) \sim \nu, s'\sim P(s,a)$, 
\begin{align} \label{eq:contraction} 
\|\Bcal f - \Bcal f'\|_\nu \le \gamma  \|f - f'\|_{P(\nu) \times \pi_{f,f'}}.
\end{align}
\end{lemma}
Although similar results are also proved in classical analyses, proving Lemma~\ref{lem:contraction} under Assumption~\ref{asm:bi_close} is more challenging. For example, a very useful property in the classical analysis is that $\EE[(\Tcal f)(s,a) - r - \gamma \max_{a'} f(s',a')|s,a]=0$, and it holds in a point-wise manner for every $(s,a)$. In our \bi analyses, however, such a handy tool is not available as we only make assumptions on the average-case properties of the functions, and their point-wise behavior is undefined. %While the proof is more challenging, working it out is beneficial as it forces us to think about which part of our assumptions is really necessary and which part is not.  
We refer the readers to Appendix~\ref{app:fqi} for how we overcome this technical difficulty.

With $f^\star$ defined in Theorem~\ref{thm:fstar}, we state the main theorem of this section, with proof deferred to Appendix~\ref{app:fqi}.
\begin{theorem} \label{thm:fqi}
Let $f_1, f_2, \ldots, f_k \in \Fcal$ be the sequence of functions obtained by FQI. Let $\epsilon$ be an universal upper bound on the error incurred in each iteration, that is, $\forall 1\le i \le k-1$,
$$
\Lcal_{\mu}(f_k; f_{k-1}) \le \min_{f \in \Fcal} \Lcal_{\mu}(f; f_{k-1}) + \epsilon.
$$
Let $\hat \pi$ be the greedy policy of $f_k$. Then
$
v^{\hat \pi} \ge v^{\pi_{f^\star}} - \frac{2}{1-\gamma}(\frac{\sqrt{C\epsilon}}{1-\gamma} + \gamma^k \Vmax).
$
\end{theorem}

\section{Further Discussions} % and Conclusions
\label{sec:conclusion}

\subsection{List of \BD Quantities} \label{sec:bd_list}
Below we provide a list of quantities that are \bxd, and discuss the source of their \bdc and connect to related literature. 

\para{$V^\pi, V^\star, Q^\pi, Q^\star$:} Their \bdc should be obvious from the earlier discussions. 

\para{$\pi^\star$, $v^\star$: } Since $\pi^\star$ is tightly bonded to $Q^\star$, its \bdc should not be surprising. Appendix~\ref{app:il} also provides an example where $\pi^\star$ changes with the boundary. The \bdc of $v^\star$ stems from that of $\pi^\star$. In fact, for any fixed $\pi$, the expected return \emph{under the initial distribution}, $v^\pi$, is \bxi, as $v^\pi$ can be estimated efficiently\footnote{Here by ``efficiently'' we mean that the error of MC policy evaluation is $O(1/\sqrt{n})$ (which is a direct consequence of Hoeffding's) without any function approximation assumptions (such as realizability), and the constant does not depend on the size of the state space (c.f.~Proposition~\ref{prop:verify}). } by Monte-Carlo policy evaluation without specifying a boundary. 

\para{$\Tcal^\pi, \Tcal$:} It should not be surprising that the Bellman update operators are \bxd, as the value functions are tightly bonded to these operators. 

\para{Bellman errors $\|f - \Tcal f\|$ and $\|f - \Tcal^\pi f\|$:} This is perhaps the most interesting case, as many algorithms in RL are designed to minimize the Bellman error. \citet[Chapter 11.6]{sutton2018reinforcement} provide several carefully constructed examples to show that Bellman error of a single candidate function is fundamentally unlearnable (even in the limit of infinite data). From the perspective of our paper, \emph{this result is obvious:} Bellman errors are \bxd so they are of course unlearnable (from natural data)!

We note, however, that the \bdc of Bellman errors does \emph{not} invalidate the Bellman error minimization approaches  \citep[e.g.,][]{antos2008learning, dai2018sbeed}, as none of those approaches directly estimate the Bellman errors due to the difficulty of the conditional expectation in $\Tcal$ and $\Tcal^\pi$ (which is precisely their source of \bdc; see related discussions in Section~\ref{sec:verify}); we refer the readers to \citet{chen2019information} for discussions on this classical difficulty of value-based RL. 

Without going deeper into the issue, let us summarize our conclusion in a way that is consistent with the view of this paper and the results in prior work: \\
(1) Bellman error is ill-defined for \emph{a single function} (this is in agreement with \citet{sutton2018reinforcement}). \\
(2) When we have \emph{a function class} that satisfies certain (verifiable) properties (e.g., Assumption~\ref{asm:bi_close}), the Bellman error of a function \emph{with repsect to the function class}  may be well-defined (this is consistent with the existing work on Bellmen error minimization).

\subsection{On State Resetting, MCTS, and Imitation Learning} \label{sec:reset}
The unlearnability of Bellman errors seemingly contradicts the fact that Bellman errors have an unbiased estimator using the double sampling trick \citep{baird1995residual}: given a data tuple $(s,a,r,s')$, if we can obtain another sample $s'' \sim P(s, a)$, where $s'$ and $s''$ are i.i.d.~conditioned on $(s, a)$, then the following estimator for Bellman error is unbiased (up to reward variance): %(assuming deterministic rewards):
$$
(f(s, a) - r - \gamma \max_{a'} f(s', a')) (f(s, a) - r - \gamma \max_{a'' } f(s'', a'')). 
$$
There is no real contradiction here, as 
double sampling requires a very special data collection protocol---essentially the ability to reset states. Such a protocol is not supported by the data generation process considered in the technical sections of this paper, and is likely only available in simulated environments. 
Furthermore,  state resetting itself is \bxd (see concrete examples in the next paragraph): when we sample $s'' \sim P(s,a)$, the distribution of $s''$ depends on how we reproduce $s$, i.e., the boundary, and the estimated Bellman error is w.r.t.~that boundary. 

%Besides residual algorithms, a
Another family of \bd algorithms is Monte-Carlo tree search \citep[MCTS; e.g.,][]{kearns2002sparse, kocsis2006bandit}. At each time step, MCTS rolls out multiple trajectories from the \emph{current state} to determine the optimal action, which also requires state resetting. Among many ways of resetting the state, one can clone the RAM configuration and reset to that (``\textit{boundary 0}'' in Figure~\ref{fig:boundary}). One can also attempt to reproduce the sequence of observations and actions from the beginning of the episode \citep[see POMCP; ][]{silver2010montea}. Both are valid state resetting operations but for different boundaries. 

The \bdc of MCTS brings questions to the popular approach of combining MCTS with function approximation. For example, \citet{guo2014deep} have trained deep neural networks to imitate the MCTS policy and its learned Q-values. In such approaches, the boundary of MCTS is often chosen to be ``boundary 0'' (i.e., cloning RAM) due to its convenience, but when the definition of state includes the pseudorandom generator, the evolution of the system becomes fully deterministic from the resetting state even if the original problem is stochastic. Since the function approximation agent often takes visual observations as perceptual inputs and cannot observe the state of the pseudorandom generator, this can create a mismatch between the demonstration policy and the capacity of the learner. In fact, we show an extreme result in Appendix~\ref{app:il} that supports the following claim:
\begin{proposition}[Informal] \label{prop:useless}
For a learner restricted by a more lossy boundary, demonstrations generated by $\pi^\star$ can be completely useless. 
\end{proposition}
In the imitation learning (IL) literature, a similar issue is known regarding algorithms that directly learn the expert's actions \citep[such as DAgger;][]{ross2011reduction}, that their performance can be degenerate if the learner's policy class cannot represent the expert policy. In fact, this is a major motivation for considering return-weighted regression IL algorithms \citep[such as AggreVaTe;][]{ross2014reinforcement}, which is believed to address this issue. Our counterexample shows that return-weighted regression is NOT the elixir for model mismatch in IL: when the expert's and the learner's boundaries are different, the demonstrations could be useless whatsoever even if the learner is allowed to run return-weighted regression. While addressing this issue is beyond the scope of this paper, we believe that our paper provides a useful conceptual framework for reasoning about such problems.

%In Section~\ref{sec:dilemma} we show that the issue of \bdc is not just conceptual puzzles and can have real consequences, especially when MCTS and value-function approximation appear together. One can further ask: When we use MCTS to provide expert demonstration for a value-based learner \citep[e.g.,][]{guo2014deep}, how should we choose the boundary (i.e., which notion of state should we reset to in MCTS)? (The same question can be also asked about the residual algorithms \citep{baird1995residual}, which minimize Bellman errors via the \emph{double sampling} trick, i.e., drawing two i.i.d.~next-states $s'$ from the same $(s,a)$.) More generally, when the learner is of limited capability in an imitation learning scenario \citep{ross2011reduction, ross2014reinforcement}, how to best design the demonstration policy? In fact, we show in Appendix~\ref{app:il} that demonstration using $\pi^\star$ for a poorly chosen boundary can be \emph{completely} useless. Answering these questions is beyond the scope of this paper, and we leave the investigation to future work. 

\subsection{Verifiability of Theoretical Assumptions}  \label{sec:verify}
%The ability to verify the correctness of theoretical assumptions is important to the development and the debugging of RL algorithms. 
As Section~\ref{sec:intro} already alluded, classical realizability-type assumptions  not only are \bxd, but also cannot be verified from naturally generated data. In fact, this claim can be formalized as below: 
\begin{proposition}\label{prop:verify}
Under the setting of Section~\ref{sec:fqi_intro}, there exists $\Fcal$ of finite cardinality and constant $\epsilon$, such that for any $\delta > 0$ and sample size $n$, the nature can choose a data generating distribution adversarially, and no algorithm can determine the realizability of $\Fcal$ up to $\epsilon$ error with success probability $\ge 1/2 + \delta$ from a finite sample of size $n$.
\end{proposition}
%Appendix~\ref{app:verify}, we provide a formal information-theoretic hardness result showing that realizability (i.e., $Q^\star \in \Fcal$) is fundamentally unverifiable. 
The complete argument and proof are provided in Appendix~\ref{app:verify}. The major difficulty is that quantities like $Q^\star$ are defined via conditional expectations ``$\EE[\,\cdot \cond s_1 =s, a_1 =a]$'', which cannot be estimated unless we can  reproduce the same state multiple times (i.e., state resetting; see Section~\ref{sec:reset}). This issue is eliminated in the \bi analyses, as the assumptions are stated using plain expectations over data (recall Claim~\ref{clm:bi}), which can be verified (up to any accuracy) via Monte-Carlo estimation. 

Of course, verifying the \bi assumptions still faces significant challenges, as the statements frequently use languages like ``$\forall f\in\Fcal$'' and ``$\forall$ admissible $\nu$'', making it computationally expensive to verify them exhaustively. We note, however, that this is likely to be the case for any strict theoretical assumptions, and practitioners often develop  heuristics under the guidance of theory to make the verification process tractable. For example, the difficulty related to ``$\forall f\in\Fcal$'' may be resolved by clever optimization techniques, and that related to ``$\forall \nu$'' may be addressed by testing the assumptions on a diverse and representative set of distributions designed with domain knowledge. We leave the design of an efficient and effective verification protocol to future work.

\subsection{Discard \BD Analyses?} We do not advocate for always insisting on  \bi analyses and this is not the intention of this paper. Rather, our purpose is to demonstrate the feasibility of \bi analyses, and to use the concrete maths to ground the discussions of the conceptual issues (which can easily go astray and become vacuous given the nature of this topic). On a related note, \bd analyses make stronger assumptions hence are mathematically easier to work with. 

\subsection{\BIC in Other Algorithms} 
The \bi version of Bellman equation for policy evaluation has appeared in \citet{jiang2017contextual} who study PAC exploration under function approximation, although they do not discuss its further implications. While our assumptions are inspired by theirs, we have to deal with additional technical difficulties due to off-policy policy optimization.% In Appendix~\ref{app:sqloss} we discuss the connections and the differences between the two papers on a concrete example. 

\subsection{``Boundary 0''} 
As we hinted in Figure~\ref{fig:boundary}, every RL problem has an equivalent reformulation with deterministic transitions, which has been formalized mathematically by \citet{ng2000pegasus}. This leads to a number of paradoxes: for example, many difficulties in RL arise due to stochastic transitions, and there are algorithms designed for deterministic systems that avoid these difficulties. Why don't we always use them since all environments are essentially deterministic? This question, among others, is discussed in Appendix~\ref{app:dbrm}.  %The formal mathematical description has been given by  and summarized in Appendix~\ref{app:dbrm}. 
In general, we find that investigating this extreme view is helpful in clarifying some of the confusions and sparks novel ways of thinking about states and optimality in RL. %, and it provides justifications for certain design choices in our theory. %The detailed discussions are long and involved, 

%\para{Residual Algorithms} 
%\citet{baird1995residual} introduced the residual algorithms that minimize Bellman error via the \emph{double sampling} trick, i.e., drawing two i.i.d.~next-states $s'$ from the same $(s,a)$. Compared to 

\newpage

%\subsubsection*{Acknowledgments}

\bibliography{bib}
\bibliographystyle{plainnat}

\clearpage
\onecolumn
\appendix

\input{proofs}

\end{document}

%% file: def.tex
\usepackage{url, hyperref}
\usepackage{amsthm, amsmath, amssymb} %amsfonts, amsmath, bm
\usepackage{enumerate, paralist}
\usepackage{comment}
\usepackage{color}
\usepackage{multirow}
\usepackage{xspace}
\usepackage{rotating}
\usepackage{graphicx}

\theoremstyle{plain}
\newtheorem{theorem}{\textbf{Theorem}}
\newtheorem{lemma}[theorem]{\textbf{Lemma}}

\newtheorem{proposition}[theorem]{Proposition}
\newtheorem{claim}[theorem]{Claim}

\newtheorem{question}{Question}

\theoremstyle{definition}
\newtheorem{definition}{Definition}
\newtheorem{assumption}{Assumption}
\newtheorem{fact}{Fact}

\renewcommand{\cite}{\citep}

\newcommand{\figref}[1]{(\ref{fig:#1})}

\newcommand{\EE}{\mathbb{E}}

\newcommand{\cond}{~\big\vert~}

\newcommand{\Indi}{\mathbb{I}}

\DeclareMathOperator*{\argmin}{arg\,min}
\DeclareMathOperator*{\argmax}{arg\,max}

\newcommand{\Fcal}{\mathcal{F}}

\newcommand{\Scal}{\mathcal{S}}

\newcommand{\Acal}{\mathcal{A}}
\newcommand{\Tcal}{\mathcal{T}}
\newcommand{\Bcal}{\mathcal{B}}
\newcommand{\Ecal}{\mathcal{E}}

\newcommand{\Rmax}{R_{\max}}
\newcommand{\Vmax}{V_{\max}}
\newcommand{\RR}{\mathbb{R}}
\newcommand{\Xcal}{\mathcal{X}}
\newcommand{\Ycal}{\mathcal{Y}}

\newcommand{\eapx}{\epsilon_{\textrm{approx}}}

\newcommand{\fh}{f_k}
\newcommand{\fo}{f_{k-1}}

\newcommand{\Lcal}{\mathcal{L}}

\newcommand{\hf}{{\hat f}}

\newcommand{\bi}{boundary-invariant\xspace}
\newcommand{\bxi}{boundary invariant\xspace}
\newcommand{\bic}{boundary invariance\xspace}
\newcommand{\BIC}{Boundary Invariance\xspace}
\newcommand{\Bi}{Boundary-invariant\xspace}

\newcommand{\BI}{Boundary-Invariant\xspace}

\newcommand{\bd}{boundary-dependent\xspace}
\newcommand{\Bd}{Boundary-dependent\xspace}
\newcommand{\BD}{Boundary-Dependent\xspace}
\newcommand{\bxd}{boundary dependent\xspace}

\newcommand{\BDC}{Boundary Dependence\xspace}
\newcommand{\bdc}{boundary dependence\xspace}

%% file: proofs.tex
\section{Proof of Proposition~\ref{prop:verify}} \label{app:verify}
To show that realizability is not verifiable in general, it suffices to show an example in contextual bandits (Section~\ref{sec:cb}). We further simplify the problem by restricting the number of actions to $1$, which becomes a standard regression problem, and $\Fcal$ is realizable if it contains the Bayes-optimal predictor. We provide an argument below, inspired by that of the No Free Lunch theorem \citep{wolpert1996lack}.

Consider a regression problem with finite feature space  $\Xcal$ and label space $\Ycal = [0, 1]$. The hypothesis class only consists of one function, $f_{1/2}$, that takes a constant value $1/2$. We will construct multiple data distributions in the form of $P_{X, Y} \in \Delta(\Xcal\times\Ycal)$, and $f_{1/2}$ is the Bayes-optimal regressor for one of them (hence realizable) but not for the others, and in the latter case realizability will be violated by a large margin. An adversary chooses the distribution in a randomized manner, and the learner draws a finite dataset $\{(X_i, Y_i)\}$ from the chosen distribution and needs to decide whether $\Fcal = \{f_{1/2}\}$ is realizable or not. We show that no learner can answer this question better than random guess when $|\Xcal|$ goes to infinity.

In all distributions, the marginal of $X$ is always uniform, and it remains to specify $P_{Y|X}$. For the realizable case, $P_{Y|X}$ is distributed as a Bernoulli random variable independent of the value of $X$. It will be convenient to refer to a data distribution by its Bayes-optimal regressor, so this distribution is labeled $f_{1/2}$.

For the remaining distributions, the label $Y$ is always a deterministic and binary function of $X$, and there are in total $2^{|\Xcal|}$ such functions. When the adversary chooses a distribution from this family, it always draws uniformly randomly, and we refer to the drawn function (and distribution) $f_{\text{rnd}}$. Note that regardless of which function is drawn, $f_{1/2}$ always violates realizability by a constantly large margin: 
$$
\EE_X[(f_{1/2}(X) - f_{\text{rnd}}(X))^2] = 1/4.
$$
So as long as we set the $\epsilon$ in the statement of Proposition~\ref{prop:verify} to be smaller than $1/4$, verifying realizability up to $\epsilon$ error means verifying it exactly in our construction. 

The adversary chooses $f_{1/2}$ with $1/2$ probability, and $f_{\text{rnd}}$ with $1/2$ probability. Since the learner only receives a finite sample, as long as there is no collision in $\{X_i\}$, there is no way to distinguish between $f_{1/2}$ and $f_{\text{rnd}}$. This is because, $\{(X_i, Y_i)\}$ can be drawn in two steps, where we first draw all the $\{X_i\}$ i.i.d.~from Unif$(\Xcal)$, and this step does not reveal any information about the identity of the distribution. The second step generates $\{Y_i\}$ conditioned on $\{X_i\}$. Assuming no collision in $\{X_i\}$, it is easy to verify that the joint distribution over $\{Y_i\}$ is i.i.d.~Bernoulli for both $f_{1/2}$ and $f_{\text{rnd}}$. Furthermore, fixing the sample size $n$, the collision probability goes to $0$ as $|\Xcal|$ increases. So as long as $|\Xcal|$ is chosen to be large enough (depending on $\delta$ and sample size $n$ in the proposition statement), we can guarantee that no algorithm can determine the realizability with a success rate greater than $1/2+\delta$.

Note that this hardness result does not apply when the learner has access to the resetting operations discussed in Section~\ref{sec:reset}, as the learner can drawn multiple $Y$'s from the same $X$ to verify if $P_{Y|X}$ is stochastic ($f_{1/2}$) or deterministic ($f_{\text{rnd}}$) and succeed with high probability.

\section{Necessity of the Squared-loss Decomposition Condition} \label{app:sqloss}
Here we provide an example showing the necessity of Eq.\eqref{eq:sqloss} in Assumption~\ref{asm:bi_cb_realizable}. In particular, if Eq.\eqref{eq:sqloss} is completely removed, the algorithm may fail to learn a valid value function in the limit of infinite data even when $\Fcal$ contains one. 

Consider the CB problem in Figure~\ref{fig:cb1}, except that both states yield a deterministic reward of $0.5$. Let $\Fcal = \{f_1, f_2\}$, where $f_1(s_A) = 1$, $f_1(s_B) = 0$, and $f_2(s_A) = f_2(s_B) = 0.6$. By Assumption~\ref{asm:bi_cb_realizable} (with Eq.\eqref{eq:sqloss} removed), $f_1$ is a valid reward function while $f_2$ is not. However,  
$\Lcal_{\mu}(f_1) = 0.25$ and $\Lcal_{\mu}(f_2) = 0.01$, and the regression algorithm will pick $f_2$ with accurate estimation of the losses. 

%\para{Further Comments} In the above example, we note that there is nothing wrong in calling $f_1$ a valid reward function (though it is counter-intuitive). In fact, if this bandit problem is a part of a larger MDP---say it appears at the end of an episodic task, and $d_0$ is the only possible distribution that can be induced over $s_A$ and $s_B$, then $f_1$ may well be part of an optimal value function, and a near-optimal policy can be learned via active exploration using the OLIVE algorithm \cite{jiang2017contextual}.\footnote{Formally, $f_1$ does not violate the validity condition in their Definition 3.} The reason that $f_1$ should not be considered as a valid reward function in the context of Section~\ref{sec:cb} is due to the batch learning setting and the squared-loss regression algorithm. So Eq.\eqref{eq:sqloss} is a condition that is specific to the setting and the algorithm, and not inherent in our \bi definition of reward/value-functions.
 
\newpage
 
\section{Proofs of Sections~\ref{sec:cb} to \ref{sec:fqi}} \label{app:fqi}
\begin{proof}[\textbf{Proof of Theorem~\ref{thm:cb}}]
Since $f^\star$ is a valid reward function (Eq.\eqref{eq:reward}), we have
\begin{align*}
&~ v^{\pi_{f^\star}} - v^{\pi_\hf} = 	\EE_{d_0 \times \pi_{f^\star}}[f^\star] - 	\EE_{d_0 \times \pi_\hf}[f^\star] \\
\le &~ \EE_{s\sim d_0}[f^\star(s,\pi_{f^\star}) - \hf(s, \pi_{f^\star}) + \hf(s, \pi_\hf) - f^\star(s,\pi_\hf)]\\
= &~ \EE_{d_0 \times \pi_{f^\star}}[f^\star - \hf] + \EE_{d_0 \times \pi_{\hf}}[\hf - f^\star].
\end{align*}
Let $\nu$ be a placeholder for either $d_0 \times \pi_\hf$ or $d_0 \times \pi_{f^\star}$. For either of the 2 terms, its square can be bounded as
\begin{align*}
&~ (\EE_{\nu}[f^\star - \hf])^2 \le \EE_{\nu}[(f^\star - \hf)^2] = \|f^\star - \hf\|_{\nu}^2 \tag{Jensen}\\
\le &~ C\cdot \|f^\star - \hf\|_{\mu}^2 
\tag{Assumption~\ref{asm:bi_cb_explore}}\\
= &~ C \cdot (\Lcal_{\mu}(\hf) - \Lcal_{\mu}(f^\star)) \tag{Eq.\eqref{eq:sqloss}} \\
= &~ C \cdot (\Lcal_{\mu}(\hf) - \min_{f \in \Fcal}\Lcal_{\mu}(f)) 
\le C \epsilon. \tag*{\qedhere}
\end{align*}
%So we conclude that $\EE_{d_0 \times \pi_{f^\star}}[f^\star] - v^{\pi_\hf} \le 2\sqrt{C\epsilon}$.
\end{proof}

\begin{proof}[\textbf{Proof of Theorem~\ref{thm:cb_robust}}]
The proof is mostly identical to that of Theorem~\ref{thm:cb} except for two differences: (1) $f^\star$ should be replaced by $Q^\star$, and (2) in the last step, we have
\begin{align}
(\EE_{\nu}[Q^\star - \hf])^2 \le &~ C(\Lcal_\mu(\hf) - \Lcal_\mu(Q^\star)) \\
= &~ C(\Lcal_\mu(\hf) - \min_{f\in \Fcal}\Lcal_\mu(f) + \eapx) \le C(\epsilon+\eapx). 
\end{align}
Plugging this into the rest of the analysis completes the proof.
\end{proof}

\begin{proof}[\textbf{Proof of Proposition~\ref{prop:bi_better}}]
Since any $C$ that satisfies Assumption~\ref{asm:cb_explore} also satisfies Assumption~\ref{asm:bi_cb_explore}, it suffices to show that Theorem~\ref{thm:cb} is tighter than Theorem~\ref{thm:cb_robust} when they use the same $C$, that is, 
\begin{align} \label{eq:bi_better_wts}
v^\star - 2\sqrt{C(\epsilon + \eapx)} \le v^{\pi_{f^\star}} - 2\sqrt{C \epsilon}.
\end{align} 
First, notice that 
\begin{align*}
v^\star - v^{\pi_{f^\star}} 
= &~ \EE_{\pi^\star}[Q^\star] - \EE_{\pi_{f^\star}}[f^\star] 
\le \EE_{\pi^\star}[Q^\star] - \EE_{\pi^\star}[f^\star] \\
\le &~ |\EE_{d_0 \times \pi^\star}[Q^\star - f^\star]| 
\le \|Q^\star - f^\star\|_{d_0 \times \pi^\star} \\
\le &~ \sqrt{C}\|Q^\star - f^\star\|_{\mu}  \tag{Assumption~\ref{asm:cb_explore}} \\
= &~ \sqrt{C \eapx}. \tag{Assumption~\ref{asm:bi_cb_realizable}: $f^\star$ minimizes squared loss in $\Fcal$} 
\end{align*} 
So now it suffices to show that $2\sqrt{C \epsilon} + \sqrt{C \eapx} \le 2\sqrt{C(\epsilon + \eapx)}$. 
Dropping $\sqrt{C}$ on both sides, the LHS becomes
\begin{align*}
2\sqrt{\epsilon} + \sqrt{\eapx} \le &~ \sqrt{8\epsilon + 2\eapx} \tag{$\frac{a+b}{2} \le \sqrt{\frac{a^2+b^2}{2}}$} \\
\le &~ \sqrt{4\epsilon + 4\eapx} \tag{the assumption that $\epsilon \le \eapx/2$}\\
= &~ 2\sqrt{\epsilon + \eapx}.
\end{align*} 
This completes the proof.
\end{proof}

\begin{proof}[\textbf{Proof of Lemma~\ref{lem:contraction}}] Let $\Bcal_{r,s'} f$ be a shorthand for $r + \gamma \max_{a'\in\Acal} f(s',a')$. 
Also recall that $\EE_{\nu}[\cdot]$ is short for $\EE_{(s,a)\sim \nu, r\sim R(s,a), s'\sim P(s,a)}[\cdot]$, and $\EE_{\nu}[f]$ for $\EE_{\nu}[f(s,a)]$. The first step is to show that 
\begin{align} \label{eq:jensen}
\|\Bcal f - \Bcal f'\|_\nu^2 \le 
\EE_{\nu}[(\Bcal_{r,s'} f - \Bcal_{r,s'} f')^2].
\end{align}
To prove this, we start with Eq.\eqref{eq:fqi_sqloss}: %Assumption~\ref{asm:bi_close} that $\Bcal f'$ is a reward function for random reward $\Bcal_{r,s'} f'$, so
\begin{align*}
0 = &~ \Lcal_{\mu}(\Bcal f; f') - \Lcal_{\mu}(\Bcal f'; f') - \|\Bcal f - \Bcal f'\|_\mu^2.\\
= &~ \EE_{\nu}[(\Bcal f - \Bcal_{r,s'} f')^2] - \EE_{\nu}[(\Bcal f' - \Bcal_{r,s'} f')^2] - \EE_{\nu}[(\Bcal f - \Bcal f')^2] \\
= &~ 2\, \EE_{\nu}[(\Bcal f' - \Bcal_{r,s'} f')(\Bcal f - \Bcal f')].
\end{align*}
%We remind the readers that $\Bcal f$ and $\Bcal f'$ above should have been $(\Bcal f)(s,a)$ and $(\Bcal f')(s,a)$ respectively, and we drop $(s,a)$ for readability. 
Now we have 
\begin{align} \label{eq:bfbf1}
\EE_{\nu}[(\Bcal f - \Bcal f')(\Bcal f' - \Bcal_{r,s'} f')] = 0,
\end{align}
and by symmetry 
\begin{align} \label{eq:bfbf2}
\EE_{\nu}[(\Bcal f' - \Bcal f)(\Bcal f - \Bcal_{r,s'} f)] = 0.
\end{align}
We are ready to prove Eq.\eqref{eq:jensen}: its RHS is
\begin{align*}
&~ \EE_{\nu}[(\Bcal_{r,s'} f - \Bcal_{r,s'} f')^2] \\
= &~\EE_{\nu}[(\Bcal_{r,s'} f - \Bcal_{r,s'} f' - \Bcal f + \Bcal f' + \Bcal f - \Bcal f')^2] \\
= &~\EE_{\nu}[(\Bcal_{r,s'} f - \Bcal_{r,s'} f' - \Bcal f + \Bcal f')^2] + \EE_{\nu}[(\Bcal f - \Bcal f')^2]  \\ 
&~ + 2\EE_{\nu}[(\Bcal f - \Bcal f')(\Bcal_{r,s'} f - \Bcal f)] + 2\EE_{\nu}[(\Bcal f - \Bcal f')(\Bcal f' - \Bcal_{r,s'} f')].
\end{align*}
The 1st term is non-negative, the 2nd term is the LHS of Eq.\eqref{eq:jensen}, and the rest two terms are $0$ according to Eq.\eqref{eq:bfbf1} and \eqref{eq:bfbf2}. So Eq.\eqref{eq:jensen} holds.

Now from the RHS of Eq.\eqref{eq:jensen}: 
\begin{align*}
&~ \EE_{\nu}[(\Bcal_{r,s'} f - \Bcal_{r,s'} f')^2] \\
= &~ \gamma^2 \EE_{(s,a)\sim \nu, s' \sim P(s,a)}[(f(s',\pi_f) - f'(s',\pi_{f'}))^2] \\
= &~ \gamma^2 \EE_{s' \sim P(\nu)}[(f(s',\pi_f) - f'(s',\pi_{f'}))^2] \\
\le &~ \gamma^2 \EE_{s' \sim P(\nu)}[(f(s', \pi_{f,f'}) - f'(s', \pi_{f,f'}))^2] \\
= &~ \gamma^2 \| f - f'\|_{P(\nu) \times \pi_{f,f'}}^2. \tag*{\qedhere}
\end{align*}
\end{proof}

\begin{proof}[\textbf{Proof of Theorem~\ref{thm:fstar}}]
Since $\Bcal f \in \Fcal$ for any $f\in\Fcal$ by our definition, we can  apply $\Bcal$ repeatedly to a function. Indeed, pick any $f\in\Fcal$, we show that for large enough $k$, $\|\Bcal^{k+1} f - \Bcal^k f\|_\nu = 0$ for any admissible $\nu$, so $\Bcal^k f$ will satisfy the definition of $f^\star$. This is because
\begin{align*}
\|\Bcal^{k+1} f - \Bcal^k f\|_\nu \le 
&~ \gamma \|\Bcal^{k} f - \Bcal^{k-1} f\|_{P(\nu)\times \pi_{\Bcal^k f, \Bcal^{k-1} f'}} \tag{Lemma~\ref{lem:contraction}} \\
\le &~ \gamma^2 \|\Bcal^{k-1} f - \Bcal^{k-2} f\|_{P(P(\nu)\times \pi_{\Bcal^k f, \Bcal^{k-1} f'}) \times \pi_{\Bcal^{k-1} f, \Bcal^{k-2} f'}} \\
\le &~ \cdots \le \gamma^k \|\Bcal f - f\|_{\square} \le \gamma^k \|\Bcal f - f\|_{\infty},
\end{align*}
where $\square$ is some admissible distribution. (Its detailed form is not important, but the reader can infer from the derivation above.) Given the boundedness of $\Fcal$, $\|\Bcal^{k+1} f - \Bcal^k f\|_\nu$ becomes arbitrarily close to $0$ for all $\nu$ uniformly as $k$ increases.  
Now for each $f\in\Fcal$, define $\Ecal(f):= \sup_{\nu} \|\Bcal f - f\|_\nu$. Since $\Fcal$ is finite, there exists a minimum non-zero value for $\Ecal(f)$, so with large enough $k$, $\sup_{\nu}\|\Bcal^{k+1} f - \Bcal^k f\|_\nu$ will be smaller than such a minimum value and must be $0$.
\end{proof}

\paragraph{Comment} In the proof of Theorem~\ref{thm:fstar} we used the fact $\Fcal$ is finite to show that $\|\Bcal^{k+1} f - \Bcal^k f\| = 0$. This is the only place in this paper that uses the finiteness of $\Fcal$. Even if $\Fcal$ is continuous, we can still use a large enough $k$ to upper bound $\|\Bcal^{k+1} f - \Bcal^k f\|$ with an arbitrarily small number.

\begin{comment}
Finally we prove Theorem~\ref{thm:fqi}. To prove it, we introduce a few further notations and helper lemmas.

\paragraph{Matrix Notations} We use $\Pi \in \RR^{|\Scal|\times |\Scal\times\Acal|}$ to denote the matrix form of a policy $\pi$, that is, $\Pi_{s, (s,a)} = \Indi[s=s'] \cdot \pi(a|s)$. Similarly $\Pi_f$ is the matrix form of $\pi_f$. Let $P^\pi \in \RR^{|\Scal|\times|\Scal|}$ be the transition matrix of the Markov chain induced by $\pi$. We will also treat $V^\pi$ and $Q^\pi$ as vectors whenenever appropriate.

\begin{lemma}
For any $f, f'$, 
$v^{\pi_f} - v^{\pi_{f'}} \le (I - \gamma P^{\pi_f})^{-1} (\Pi_f - \Pi_{f'}) $
\end{lemma}
\end{comment}

\begin{proof}[\textbf{Proof of Theorem~\ref{thm:fqi}}]
We first show that $f^\star$ is a value-function of $\pi_{f^\star}$  on any admissible $\nu$. The easiest way to prove this is to introduce the classical (boundary-dependent) notion of $Q^{\pi_{f^\star}}$ as a bridge. Note that $\forall \nu$ we always have
$$ \textstyle
\EE_{\nu}[Q^{\pi_{f^\star}}] = \EE[\sum_{t=1}^\infty \gamma^{t-1} r_t \cond (s_1, a_1) \sim \nu, a_{2:\infty} \sim \pi_{f^\star}].
$$
So it suffices to show that $\forall \nu$, $\EE_{\nu}[Q^{\pi_{f^\star}}] = \EE_{\nu}[f^\star]$.
We prove this using (a slight variant of) the value difference decomposition lemma \citep[Lemma 1]{jiang2017contextual}:
\begin{align*}
\textstyle\EE_{\nu}[f^\star] - \EE_{\nu}[Q^{\pi_{f^\star}}]   =\sum_{t=1}^\infty \gamma^{t-1} \EE_{(s,a) \sim d_{\nu, t}^{\pi}}[f^\star(s,a) - r - \gamma \max_{a'\in \Acal} f^\star(s', a')].
\end{align*}
Here with a slight abuse of notation we use $d_{\nu, t}^\pi$ to denote the distribution over $(s_t,a_t)$ induced by $(s_1, a_1) \sim \nu$, $a_{2:t-1} \sim \pi$, and $\pi = \pi_{f^\star}$. For each term on the RHS,
\begin{align*}
&~ \EE_{(s,a) \sim d_{\nu, t}^{\pi}}[f^\star(s,a) - r - \gamma \max_{a'\in \Acal} f^\star(s', a')] \\
= &~ \EE_{(s,a) \sim d_{\nu, t}^{\pi}}[f^\star(s,a) - (\Bcal f^\star)(s,a) + (\Bcal f^\star)(s,a) - r - \gamma \max_{a'\in \Acal} f^\star(s', a')] \\
= &~ \EE_{d_{\nu, t}^{\pi}}[f^\star - \Bcal f^\star] + \EE_{(s,a) \sim d_{\nu, t}^{\pi}}[(\Bcal f^\star)(s,a) - r - \gamma \max_{a'\in \Acal} f^\star(s', a')].
\end{align*}
The second term is $0$ because by the definition of $\Bcal$ (Assumption~\ref{asm:bi_close}): $\Bcal f^\star$ is a reward function for random reward $r + \gamma \max_{a'\in \Acal} f^\star(s', a')$ under any admissible distribution, including $d_{\nu,t}^\pi$. The first term is also $0$ because
\begin{align*}
|\EE_{d_{\nu, t}^{\pi}}[f^\star - \Bcal f^\star]|^2
\le \EE_{d_{\nu, t}^{\pi}}[(f^\star - \Bcal f^\star)^2] = 0. \tag{Theorem~\ref{thm:fstar}}
\end{align*}
	
Now
\begin{align}\label{eq:decompose}
v^{\pi_{f^\star}} - v^{\hat \pi} 
= &~ \sum_{t=1}^\infty \gamma^{t-1} \EE_{(s,a)\sim d^{\hat \pi}_t} [Q^{\pi_{f^\star}}(s, \pi_{f^\star}) - Q^{\pi_{f^\star}}(s,a)] \nonumber \tag{see e.g., \citet[Lemma 6.1]{kakade2002approximately}} \\
= &~ \sum_{t=1}^\infty \gamma^{t-1} \EE_{(s,a)\sim d^{\hat \pi}_t} [f^\star(s, \pi_{f^\star}) - f_k(s,\pi_{f^\star}) + f_k(s, a) - f^\star(s,a)] \nonumber \tag{$\EE_{\nu}[Q^{\pi_{f^\star}}] = \EE_{\nu}[f^\star]$}\\
\le &~ \sum_{t=1}^\infty \gamma^{t-1} \left(\|\fh - f^\star \|_{\eta_t^{\hat \pi} \times \pi_{f^\star}} + \| \fh - f^\star\|_{d_t^{\hat \pi}}\right), \label{eq:fk_minus_fstar}
\end{align}
where $\eta_t^{\hat \pi}$ is the marginal of $d_t^{\hat \pi}$ on states. Since both $d_t^{\hat \pi}$ and $\eta_t^{\hat \pi} \times \pi_{f^\star}$ are admissible distributions, it suffices to upper-bound $\|f^\star - f_k\|$ on any admissible distribution $\nu$. In particular,
\begin{align*}
\|\fh - f^\star\|_{\nu} 
= &~ \|f_k - \Bcal \fo + \Bcal \fo - f^\star\|_{\nu} \\
\le &~ \|\fh - \Bcal \fo\|_{\nu} + \|\Bcal \fo - \Bcal f^\star\|_{\nu} \\
\le &~ \|\fh - \Bcal \fo\|_{\nu} + \gamma \| \fo - f^\star\|_{P(\nu) \times \pi_{\fo, f^\star}}. \tag{Lemma~\ref{lem:contraction}}
\end{align*}
Note that the second term is also w.r.t.~an admissible distribution, so the inequalities can be expanded all the way to $\|f_0 - f^\star\|$. For the first term, 
\begin{align*}
&~ \|\fh - \Bcal \fo\|_{\nu} \le \sqrt{C} \|\fh - \Bcal \fo\|_{\mu} \\
= &~ \sqrt{C} \sqrt{\Lcal_{\mu}(\fh; \fo) - \Lcal_{\mu}(\Bcal \fo; \fo)} \le \sqrt{C\epsilon}. \tag{Eq.\eqref{eq:sqloss}}
\end{align*}
Altogether we have on any admissible $\nu$, 
$$
\|f_k - f^\star\|_{\nu} \le \frac{\sqrt{C\epsilon}}{1-\gamma} + \gamma^k \Vmax.
$$
The proof is completed by applying this bound to Eq.\eqref{eq:fk_minus_fstar}.
\end{proof}

\section{Boundary 0} 
\label{app:dbrm}
There is an extreme choice of the boundary for every RL problem, where the environment part always has \emph{deterministic} transition dynamics and a possibly stochastic initial state distribution. The construction has been given by \citet{ng2000pegasus}, and we briefly describe the idea here: All random transitions can be viewed as a deterministic transition function that takes an additional input, that is, there always exists a deterministic function $T$, such that
$$
s \sim P(s,a)  ~~\Leftrightarrow~~ s = T(s,a, \sigma), 
$$
where $\sigma$ is a random variable from some suitable distribution (e.g., Unif($[0,1]$)). Now we augment the state representation of the MDP to include all the $\sigma$'s that we ever need to use in an episode, and generate them at the beginning of an episode (hence random initial state) so that all later transitions become deterministic. Of course, any realistic agent should not be able to observe the $\sigma$'s, and this restriction is reflected by the fact that any $f \in \Fcal$ cannot depend on $\sigma$. If the environment is simulated on a computer, then ``boundary 0'' in Figure~\ref{fig:boundary} is a good approximation of this situation, where the pseudorandom generator plays the role of $\sigma$'s. 

Below we discuss a few topics in the context of this construction. 

\para{Algorithms for deterministic environments} Many difficulties in RL arise due to stochastic transitions, and there are algorithms for deterministic environments that avoid these difficulties. For example, learning with bootstrapped target (e.g., temporal difference, $Q$-learning) can diverge under function approximation  \citep{van1994feature, gordon1995stable, tsitsiklis1997analysis}, but if the environment is deterministic, one can optimize $\EE[(f(s,a) - r- \gamma \max_{a'\in\Acal}f(s',a'))^2]$ under an exploratory distribution to learn the $Q^\star$, and the process is always convergent and enjoys superior theoretical properties.\footnote{This is a special case of the residual algorithms introduced by \citet{baird1995residual}. The residual algorithms require double sampling (sampling two i.i.d.~$s'$ from the same $(s,a)$), which is not needed in deterministic environments.} Now we just argued that all RL problems can be viewed as deterministic; why isn't everyone using the above algorithm instead of TD/$Q$-learning?

The reason is that the algorithm tries to learn the $Q^\star$ of the deterministic environment defined by boundary 0, essentially competing  with an omnipotent agent that has precise knowledge of the outcomes of all the random events ahead of time. Since the actual function approximator does not have access to such information, realizability will be severely impaired. 

\para{On the role of $d_0$} 
In Section~\ref{sec:prelim} we specify an initial state distribution $d_0$ for the MDP. While this is common in modelling episodic tasks, a reasonable question to ask is what if the agent can start with any state (distribution) of its own choice. Our theory can actually handle this case pretty easily: we simply need to add all possible initial distributions and the downstream distributions induced by them into the definition of admissible distributions (Definition~\ref{def:admissible}). 

Without this modification, the theory will break down, and an obvious counterexample comes from the boundary 0 construction: %When the $\sigma$'s are included as part of the MDP's state, the only natural initial state distributions are the ones that draw the $\sigma$'s i.i.d.~from the predetermined distributions. 
If the agent is allowed to start from an arbitrary state deterministically, there will be no randomness in the trajectory, and the data sampled from such an initial state does not truthfully reflect the stochastic dynamics of the environment. 

\para{Should $\mu$ also be admissible?}
From the counterexample above, we see that non-admissible distributions can be problematic. This leads to the following question: Shouldn't $\mu$ be admissible, since it describes the data on which we run the learning algorithm? Interestingly, in Section~\ref{sec:fqi} we did not make such an assumption and the analysis still went through. We do not have an intuitive answer as to why, and the only explanation is that Assumption~\ref{asm:bi_explore} prevented the degenerate scenarios from happening. 

\section{$\pi^\star$ Can be Useless in Imitation Learning (Proposition~\ref{prop:useless})} \label{app:il}
Here we show an example where $\pi^\star$ (for a poorly chosen boundary) is useless for the purpose of imitation learning. Consider any episodic RL problem that has no intermediate rewards and the terminal reward $r$ is Bernoulli distributed and the mean lies in $[0.5, 1]$. We transform the problem in a way indistinguishable for the learner as follows: Whenever a random reward $r\sim \text{Bernoulli}(p)$ is given at the end, we replace that with a random transition to two states, each of which has two actions: in state $s_{A}$, action  $a_A$ yields a deterministic reward of $+1$ and $a_B$ yields $0$, and in state $s_B$, action $a_B$ yields $+1$ and $a_A$ yields $0$. When the random reward in the original problem has mean $p \in [0.5, 1]$, the transition distribution over $s_A$ and $s_B$ in the transformed problem is $p$ and $1-p$, respectively. %When $p=0.5$ in the original problem, the transition distribution to $s_A$ and $s_B$ is $(0.5, 0.5)$; when $p=1$ in the original problem, the distribution is $(1, 0)$. 
The identity information of $s_A$ and $s_B$ is not available to the agent (e.g., the function approximator treats the two states equivalently), so the transformed problem is completely equivalent to the original problem, except that the agent should always take $a_A$ in $s_A$ or $s_B$.\footnote{While the transformed problem is partially observable to the learner, giving the history information to the learner's state representation still does not help it leverage the demonstration, and we can remove partial observability by having multiple pairs of $s_A$ and $s_B$, each of which can be only visited with a fixed Bernoulli distribution.}

Suppose that we generate demonstration data from $\pi^\star$ for the transformed problem and use an imitation learning algorithm to train an agent. %Let us inspect the behavior of $\pi^\star$ in the transformed problem. 
Since $\pi^\star$ can distinguish between $s_A$ and $s_B$, it will take $a_A$ or $a_B$ depending on the observed identity and always achieve a terminal reward of $1$. In the original problem, the agent is in general supposed to take actions to maximize the value of $p$, but $\pi^\star$ in the transformed problem has no incentive to do so and can take arbitrary actions before reaching $s_A$ or $s_B$, making the demonstrations useless to the bounded agent. 